\documentclass[10pt, conference, compsocconf]{IEEEtran}
\usepackage{booktabs} 
\usepackage{url}
\usepackage{tabulary}
\usepackage{dcolumn}
\usepackage{multirow}
\usepackage{array}
\usepackage{amsmath}
\usepackage{amsfonts}
\newcolumntype{L}[1]{>{\raggedright\let\newline\\\arraybackslash\hspace{0pt}}m{#1}}
\newcolumntype{C}[1]{>{\centering\let\newline\\\arraybackslash\hspace{0pt}}m{#1}}
\newcolumntype{R}[1]{>{\raggedleft\let\newline\\\arraybackslash\hspace{0pt}}m{#1}}

\newcommand\Tstrut{\rule{0pt}{2.6ex}}         

\begin{document}
%
\title{Investigating Extensions to Random Walk Based Graph Embedding}


\author{\IEEEauthorblockN{J\"{o}rg Schl\"{o}tterer, Fatemeh Salehi Rizi, Michael Granitzer}
\IEEEauthorblockA{
University of Passau\\
Passau, Germany\\
Email: \{firstname.lastname\}@uni-passau.de}
\and
\IEEEauthorblockN{Martin Wehking}
\IEEEauthorblockA{Technical University of Munich\\
Munich, Germany\\
Email: ge49bes@mytum.de}
}

\maketitle

\begin{abstract}
Graph embedding has recently gained momentum in the research community, in particular after the introduction of random walk and neural network based approaches.
However, most of the embedding approaches focus on representing the local neighborhood of nodes and fail to capture the global graph structure, i.e. to retain the relations to distant nodes.
To counter that problem, we propose a novel extension to random walk based graph embedding, which removes a percentage of least frequent nodes from the walks at different levels.
By this removal, we simulate farther distant nodes to reside in the close neighborhood of a node and hence explicitly represent their connection.
Besides the common evaluation tasks for graph embeddings, such as node classification and link prediction, we evaluate and compare our approach against related methods on shortest path approximation.
The results indicate, that extensions to random walk based methods (including our own) improve the predictive performance only slightly - if at all.
\end{abstract}


\section{Introduction}
\label{sec:intro}
Graph analysis involves predictions over nodes, edges and further network properties, such as for example shortest paths.
A prominent example of predictions over nodes is node classification, i.e. predicting the label(s) of a node.
In a social network, we might for example predict the interests of a user or the community, this user belongs to.
Analogously, link prediction aims to identify, whether a connecting edge should exist between a pair of nodes.
In a social network for instance, link prediction can be used to discover novel connections, which are most likely to be established, i.e. users making friends.
Finding the shortest path is for example relevant in a road network in order to find the best route from A to B. 
In this case, the exact computation is easily possible, but computation costs are high, therefore a faster approximation by machine learning can be desirable.
The inherently predictive nature of the other two examples (node classification and link prediction) is in favor for a machine learning algorithm by itself.

Graph embedding aims to find meaningful feature representations (embeddings) of nodes, edges or even whole (sub-)graphs to be used as input in the aforementioned downstream machine learning tasks. 
Instead of laborious hand-engineering, those feature representations can be learned by solving an optimization problem~\cite{bengio2013representation}.
In this paper, we focus on unsupervised representation learning, i.e. learning task-independent features by defining an objective independent of the downstream prediction task.
The flexibility in the definition of the objective in the unsupervised setting allows to define computationally efficient feature learning mechanisms and the representations to be used across several tasks.

A key aspect in feature learning is to reduce the dimensionality of the original feature space in a way, that the relevant information is still retained, while noise is eliminated.
Typically, methods that account for special properties of the data perform better than general dimensionality reduction methods, such as for example PCA~\cite{jolliffe1986}.
A common graph-specific objective is to preserve the local neighborhood of a node, when learning feature representations for nodes.
However, when optimizing for the local neighborhood, the global structure might be lost in the feature representations.
Several extensions to local optimization, which aim to retain the global structure have been proposed, in particular for random walk based methods (see the next section~\ref{sec:rel-work} for details).

We propose a random walk based algorithm for learning feature representations of nodes in a network, denoted as HALK, \textbf{H}ierarchical random w\textbf{ALK}.
HALK optimizes a graph-based objective function motivated by prior work in the domain of natural language processing~\cite{mikolov2013} via stochastic gradient descent. 
The feature representations maximize the likelihood of preserving the local network neighborhood of nodes in a d-dimensional features space, while still accounting for the global neighborhood. 
We combine the feature representations of pairs of nodes using simple binary operators to arrive at feature representations for edges.
Our key contributions are as follows:
\begin{itemize}
	\item We propose a modification of the random walk sampling of existing approaches that removes a fraction of the least frequent nodes from the original walks at different levels. This moves far distant nodes closer together. By gradually increasing the amount of nodes kept until we arrive at the original walk, we represent the graph at different levels of detail. 
	\item We evaluate and compare our approach to related methods on the tasks of node classification, link prediction and shortest path approximation, providing new insights in terms of the properties captured by the different methods.\footnote{The evaluation code including methods to create embeddings for all compared approaches is available in the project repository:\\ \url{https://doi.org/10.5281/zenodo.2822192}}
\end{itemize}
\noindent The rest of the paper is structured as follows. 
In the upcoming section~\ref{sec:rel-work} we briefly survey related work in feature learning for networks with an emphasis on methods that aim to incorporate global graph structure. We present our approach and technical details in section~\ref{sec:approach}. 
In section~\ref{sec:eval}, we empirically evaluate our approach and the most closely related methods on prediction tasks over nodes, edges and shortest paths on various real-world data sets. We finally conclude and provide an outlook on possible future directions in section~\ref{sec:conclusion}.

\section{Related Work}
\label{sec:rel-work}

Graph embeddings can be obtained by either applying a general dimensionality reduction algorithm or by methods that are specifically tailored towards network-specific properties.
A wide variety has been proposed in the literature (c.f. Goyal and Ferrara~\cite{goyal2017} for a survey). 
Among the classical methods are Principal Component Analysis (PCA)~\cite{jolliffe1986}, Linear Discriminant Analysis (LDA)~\cite{martinez2001}, ISOMAP~\cite{tenenbaum2000}, Multidimensional Scaling (MDS)~\cite{kruskal1978}, LLE~\cite{roweis2000} and Laplacian Eigenmaps~\cite{belkin2002} (c.f. Yan et al.~\cite{yan2007} for a survey). Most of these methods typically rely on solving eigen decomposition and the complexity is at least quadratic in the number of nodes, which makes them inefficient to handle large-scale networks.

The advent of Word2Vec~\cite{mikolov2013} in the natural language processing domain, which places words that appear in similar contexts closely together in the embedding space, recently gave rise to neural network based methods.
One of the first methods in this line is DeepWalk~\cite{perozzi2014}, which samples random walks from the graph and treats them as sentence equivalents. 
Given the representation of a node in the embedding space, DeepWalk approximates the conditional probability of another node in the network being close by optimizing for high probabilities of nodes in the neighborhood. Neighborhood is defined by sliding a window across the sampled random walk and considering all the nodes within this window as context for the center node of the window (or neighborhood respectively).
Thereby, nodes sharing a similar neighborhood, tend to have a similar representation in the embedding space. Similar to Word2Vec implicitly factorizing a matrix of word co-occurrences~\cite{pennington2014,Levy2014}, DeepWalk has been shown to factorize a matrix of node transition probabilities~\cite{Yang2015}.

Node2Vec~\cite{Grover2016} extends DeepWalk by introducing parameters to control the random walk behaviour, aiming to discover not only neighborhood similarities (homophily) but also structural roles of nodes (structural equivalence). 
At the most extreme parameter choices, Node2Vec employs breadth-first or depth-first sampling, exploring the close-by neighborhood or nodes that are far apart in the network.

LINE~\cite{Tang2015} explicitly optimizes the embeddings to capture first- and second-order proximity, by training separate embeddings for them, which are finally concatenated. 
First-order proximity is given by explicit connections between nodes, while second-order proximity is given by comparing the nodes' neighborhoods. 
Liu et. al build on LINE and present a method that is capable to embed large-scale graphs distributively in a streaming fashion~\cite{liu2018}.

Instead of approximating the k-order proximity matrix, as DeepWalk does, GraRep~\cite{cao2015} calculates it accurately, at the cost of increased complexity. 
Yang et al.~\cite{yang2017} alleviate this problem by using information from lower order proximity matrices.
The authors of HOPE~\cite{ou2016} experimented with different similarity measures, such as Katz Index, Rooted Page Rank and Adamic-Adar.

HARP~\cite{chen2018} and Walklets~\cite{perozzi2017} address capturing higher-order proximity by adapting the random walk strategy. While HARP coarsens the graph and learns representations via hierarchically collapsed graphs, Walklets skips over steps in the random walks. Compared to DeepWalk or Node2Vec, both HARP and Walklets exhibit additional complexity, as the node representations are learned on multiple levels - The collapsing level in HARP and the skip level in Walklets.

Besides most of these approaches utilizing shallow neural network architectures to learn the feature representations, deep architectures have been proposed as well, aiming at capturing non-linearity in the graphs. SDNE~\cite{wang2016} and DNGR~\cite{cao2016} utilize autoencoders, GCN~\cite{kipf2017} defines a convolution operator on the graph.
Further, methods that incorporate additional information, such as particular graph properties, e.g. communities~\cite{Wang2017,Keikha2018} or node attributes~\cite{Yang2015,Huang2017} have been proposed. 
While random walk based methods in principle can incorporate the direction of edges during the random walk, this asymmetry is not encoded in the final embeddings. 
Khosla et al.~\cite{khosla2018} proposed an approach to maintain the different roles of nodes, according to the direction of edges.

The focus of this paper is on methods applied to the raw graph and the most closely related approaches are centered around random walks, i.e. DeepWalk, Node2Vec, HARP and Walklets. 
Experimental results reported in different papers are often hard to compare, due to varying experimental setups, evaluation metrics or datasets. 
Nevertheless, from our experience, the random walk methods deliver state of the art performance in tasks such as node classification and can compete with even far more complex models.
We will provide more details on the aforementioned random walk based methods in the upcoming section~\ref{sec:approach} when we describe our approach and also compare our method against them in section~\ref{sec:eval}.

\section{Approach}
\label{sec:approach}

\subsection{Problem Definition}
\label{ssec:problem-definiton}
Let $G=(V,E)$ denote the graph, where $V$ is the set of nodes and $E$ the set of edges, $E \subseteq (V \times V)$. 
The goal is to find an embedding for the nodes (or equivalently for the edges) 
$\phi : v \in V \rightarrow \mathbb{R}^{|V| \times d}$, 
where $d << |V|$. We want to embed the nodes into a lower-dimensional, real-valued space, while still retaining as much information as possible as in the original space.

\subsection{Feature Learning}
\label{ssec:feature-learning}
To learn the mapping defined above, we seek to maximize the following objective:
\begin{equation}
\sum_{u \in V} log P(v \in V^+|\phi(u))
\label{eq:objective}
\end{equation}
That is, we aim to maximize the log-probability of observing a node $v$ that resides in the context (denoted as $V^+$) of $u$, conditioned on its feature representation $\phi(u)$. In other words, given a node in the graph, we aim to maximize the probability of observing nodes close-by. Predicting nodes in the context is also know as the Skip-gram model (with words as the equivalent to nodes in the original model)~\cite{mikolov2013}.

We model the conditional probability of every source-context node pair as a softmax unit parametrized by a dot product of the nodes' feature representations:
\begin{equation}
P(v^+|\phi(u)) = \frac{exp(\langle \phi^\prime(v^+), \phi(u) \rangle)}{\sum_{v \in V} exp(\langle \phi^\prime(v), \phi(u) \rangle)}
\label{eq:softmax}
\end{equation}
where $\langle \cdot \rangle$ is the dot product and $\phi^\prime$ is a similar mapping as $\phi$, often referred to as projection (opposed to embedding). Technically, this is implemented by a shallow neural network, with a linear hidden layer, where the embeddings are the weights between the input layer and the hidden layer and the projections are the weights between the hidden layer and the output layer.
However, the normalization in the denominator is costly to compute and therefore, we replace $P(v^+|\phi(u))$ by negative sampling, as proposed by Mikolov et al.~\cite{mikolov2013}:
\begin{equation}
\small
log \ \sigma(\langle \phi^\prime(v^+), \phi(u) \rangle) + \sum_{k=1}^{K} \mathbb{E}_{v_k \sim P_n(v)}[log \ \sigma (-\langle \phi^\prime(v_k), \phi(u) \rangle]
\label{eq:neg-sampling}
\end{equation}
where $\sigma(x) = \frac{1}{1+exp(-x)}$ is the logistic sigmoid function, $K$ is the number of negative samples, drawn from the noise distribution $P_n(v)$, which corresponds to the frequency of nodes in the random walks (see next section for details on the random walks). 
Replacing $P(v^+|\phi(u))$ by negative sampling is possible, as we are not interested in the actual probability, but a good representation $\phi(u)$. 
The negative sampling objective is to distinguish between nodes in the context of $u$ and nodes not in the context (negative samples $v_k$). 

\subsection{Notion of Context}
\label{ssec:context}
As previously stated, we aim to maximize the probability to observe nodes that reside in the context of a node, conditioned on its feature representation. 
Up to here, we did not precisely define this context, which we will do now. 
To obtain the context, we first sample a set of truncated random walks from the graph. 
We then move a sliding window across these walks. 
The center of the sliding window is the node of interest and its spread to the left and right define the context of this node. 
The left and right spread are equal and defined by a parameter called \emph{window size}.
Each pair within this sliding window (center node and left or right spread) makes a pair for maximizing the source-context probability.
In order to emphasize the stronger connection to immediate neighbors, the window size is randomly reduced to a smaller value, giving less weight to farther distant nodes.

\subsection{Modification of Random Walk Strategy}
\label{ssec:modification}
The modification we introduce in this paper is applied after sampling the random walks.
From the original walks, we remove a fraction of least frequent nodes at different levels. 
That is, we only keep the most frequent nodes in the walks, starting at a small fraction (e.g. 10\%) and increase the retained fraction until we arrive back at the original walks.
We initialize the feature representations of \emph{all} nodes randomly, then we start training the representations of the \emph{most frequent} nodes, using the first level of reduced walks. 
At the next level, we incorporate a larger fraction of nodes, update the already trained representations and train representations for the newly added nodes.
We repeat this procedure, until we have trained representations for all nodes.

The intuition to train the most frequent nodes only in the beginning is to establish artificial connections between the most relevant nodes in the graph (hubs). 
With these artificial connections, we initialize the mapping in the embedding space by defining (close) relations between those hubs. 
The embeddings are then updated on more fine-grained levels by adding more and more nodes.

\subsection{Key Differences of Related Methods to DeepWalk}
\label{ssec:other}

In Node2Vec, the random walk sampling is parametrized, such that walks can be controlled to explore the local neighborhood or to walk further away from the original node. 
	The most extreme parameter choices, resemble Breadth-First or Depth-First sampling.

The modification of Walklets is applied after sampling the random walks. 
Nodes in the walks are skipped at different levels, from 0 to $k$. 
For each level, a separate model is trained, resulting in an overall dimensionality of the embeddings as dimensionality of a single model multiplied by the number of levels. 
These models are then combined into a single model with desired dimensionality (usually equal to the dimensionality of one of the previous single models) via PCA. 
Skipping nodes can be seen as adding artificial edges between nodes, e.g. skipping one node in the walk [a,b,c] would result in [a,c].

HARP collapses the graph at different levels before sampling the walks. Training the embeddings starts at the farthest collapsed graph, i.e. the most coarse graph, training only representations for nodes available at this level. The graph is populated back with more and more nodes on each level, until walks are sampled from the original graph. During this procedure, representations of nodes from previous levels are updated, while those for newly added nodes in the current level are initially trained.

Our approach can be seen as a kind of combination between HARP and Walklets, as training on the most frequent nodes first can be seen as collapsing the graph to its hubs, similar to HARP. 
In terms of artificial connections, our approach is similar to Walklets, as we introduce them between hubs.
Walklets adds artificial connections between every pair of nodes in the random walks, which occurs within $k$ steps in the walk, where $k$ is defined by the skip level.

\section{Evaluation}
\label{sec:eval}

We start this section with the introduction of the different evaluation tasks carried out, followed by a brief description of the datasets, we used throughout the experiments and a description of the different methods' (hyper-)parameters.
Before finally presenting the evaluation of embeddings on different tasks and datasets, we first present reproduction results from the evaluation of the methods, we compare against in section~\ref{ssec:reproduction}.

The performance of the embeddings is measured by three different prediction tasks.
The first task is label classification, i.e. we use node embeddings to predict the label of a particular node.
The second task is link prediction, in which we combine the embeddings of two nodes to an edge representation between them, in order to predict the existence of that edge in the graph.
In the last task, we use the embeddings of two nodes to predict their distance in terms of the number of edges on the shortest path between them.
%

Table~\ref{tab:fre} presents the basic statistics of the datasets used throughout the evaluation. Cora and Citeseer are citation networks, in which the class indicates the research domain (single label). 
BlogCatalog is a social network, in which edges represent friendship among bloggers and the classes represent topics a blogger is interested in (multi label).
Facebook and Youtube are also social network datasets, with edges representing friendship.

\begin{table}[t]
\center
  \caption{Datasets used in our experiments}
  \label{tab:fre}
  %

  \begin{tabular}{lllc}
    \toprule
		Dataset & \#Vertices & \#Edges & \#Classes \\
		\midrule
		Cora & 2,708 & 5,429 & 7 \\
		BlogCatalog & 10,312 & 333,983 & 39 \\
		CiteSeer & 3,312 & 4,732 & 6 \\
		Facebook & 4,039 & 88,234 & - \\
		Youtube & 1,134,890 & 2,987,624 & - \\  
  \bottomrule
\end{tabular}
\end{table}

\noindent For each graph embedding method we have to select parameters that ensure a fair comparison. 
Common parameters are the number of random walks 
$\mathbb{\mu}$, the walk length \textit{t}, window size \textit{w}, initial learning rate $\mathbb{\alpha}$, final learning rate $\mathbb{\alpha}_{min}$, the representation size \textit{d}, the amount of negative samples and the number of iterations.
Further parameters originating from Word2Vec~\cite{mikolov2013} are the sample threshold, controlling the amount of high frequency nodes that are randomly downsampled and the minimum count, ignoring nodes that occur less often in the walks than this threshold. 

The following feature learning methods have additional individual parameters:\\
\textbf{Node2Vec} extends DeepWalk by introducing 2 parameters \textit{p} and \textit{q} to control the random walk behavior. \textit{p} contributes to a depth-first search and \textit{q} to a breadth-first search like neighborhood exploration.
\\
\textbf{HARP} applies two different kinds of graph collapsing schemes. 
The graph is collapsed until only a determined number of nodes is left. 
By default, the graph is collapsed until no further collapsing is possible (i.e., the graph would consist of a single node only) and we did not deviate from this default in any evaluation task.
\\
\textbf{HALK} needs as additional parameters the percentage of the most frequent nodes that are kept for each random walk pruning level and the number of training iterations per level.
Individual learning rates per level are possible, but we used the same learning rate across all levels.
\\
\textbf{Walklets} skips nodes in a random walk. The number of skipped nodes is determined by the skip window-size $\mathbb{\pi}$ (not to be confused with \textit{w}, the window-size of the SkipGram-Model).


\subsection{Reproduction of Results}
\label{ssec:reproduction}
In order to guarantee a meaningful choice of parameters throughout the evaluation carried out in this paper and a correct implementation of the methods we compare against, we first tried to reproduce experimental results from the other papers. 
As each of these papers reports a score for node classification on BlogCatalog with a training fraction of 50\%, we selected that setup for reproduction.
We tried to stick as close as possible to the original paper, by first collecting the parameter settings as described in the papers. If we could not find a parameter value in the paper, we tried to obtain it from the source code, if available. Otherwise we selected a reasonable value according to our experience.

\noindent \textbf{Deepwalk} We derived the following parameter settings reported originally by Perozzi et al. \cite{perozzi2014}: number of random walks $\mathbb{\mu}$ = 80, window size \textit{w} = 10 and dimensionality \textit{d} = 128. 
According to the source code of Deepwalk, walk length \textit{t} = 40, intial learning rate $\mathbb{\alpha}$ = 0.025, minimal learning rate $\mathbb{\alpha}_{min}$ = 0.0001, the number of iterations is 5, the number of negative samples is 5,  the sample threshold is 0.1 and minimum count = 0.
\\
\textbf{Node2Vec} As reported by Grover et al. \cite{Grover2016}, we set  $\mathbb{\mu}$ = 10, \textit{t} = 80, \textit{w} = 10 and \textit{d} = 128, \textit{p} = 0.25 and \textit{q} = 0.25. 
We set set $\mathbb{\alpha}$ = 0.025 and $\mathbb{\alpha}_{min}$ = 0.0001, derived from the source code.
The authors report that the Skip-Gram model's number of iterations is 1. 
This resulted in worse results that were strongly below the reported ones in our evaluation. Increasing the number of iterations to 5, we were able to achieve comparable results
The other common parameters were exactly set as in Deepwalk as indicated by the source code. Additionally, we investigated the influence of the parameters \textit{p} and \textit{q} on the result. Therefore, we also ran Node2Vec with parameters \textit{p} =  \textit{q} = 1 on BlogCatalog.
\\
\textbf{Walklets} We used the parameters for the reproduction reported by Perozzi et al. \cite{perozzi2017}. We set $\mathbb{\mu}$ = 1000, \textit{t} = 11, skip window-size $\pi$ = 2 and \textit{d} = 128.  The following parameters were not reported but set by us as follows: \textit{w} = 10,  $\mathbb{\alpha}$ = 0.025 , $\mathbb{\alpha}_{min}$ = 0.001 and number of iterations = 5. We set the other common parameters equal to those of Deepwalk.
In case that the number of dimensions exceeded 128, we also used PCA for dimensionality reduction to reduce the number of dimensions to 128.
\\
\textbf{HARP} The following parameters were reported by Chen et al. \cite{chen2018}:
$\mathbb{\mu}$ = 40, \textit{t} = 10, \textit{w} = 10, \textit{d} = 128, $\mathbb{\alpha}$ = 0.025 and $\mathbb{\alpha}_{min}$ = 0.001.  
The number of iterations is not reported and set to 1 by us as we derived this setting from the code and did not achieve better scores with a higher amount of iterations. 
This behaviour can be explained by HARP's graph coarsening: Training on several levels (24 for BlogCatalog) of the coarsened graphs effectively results in several iterations.
We reused the previous common parameters again.

According to the source code, HARP and DeepWalk remove self-loops from the graph.
We applied this pre-processing step and removed isolates (nodes without any edge) afterwards.
Self-loops and isolates were only present in the Citeseer dataset, which had 3279 nodes after pre-processing.
All authors report that they use a one-vs-rest logistic regression classifier with L2 regularization for their node classification task and we replicate this setup.
Since BlogCatalog is multi-label, we first obtain the number of actual labels to predict for each sample from the test set. 
Then we predict the k most probable classes, where k is the number of labels to predict. This is a common choice in the evaluation setup of the reproduced methods.
All methods report the Macro-F1 score, except for Walklets, reporting Micro-F1, which we follow in the reproduction.
Each method uses a fraction of 50\% for training. Scores are averaged over 10 random splits and we make sure that every method sees the same splits for training and test.
\begin{table}
	\center
  \caption{Reproduction results, original score in brackets. Macro-F1 reported for all methods except Walklets (Micro)}
  \label{tab:rpsc}
  \begin{tabular}{ll}
    \toprule
    Algorithm/Dataset&BlogCatalog\\
		\midrule
    Deepwalk &27.84 (27.30) \\
    HARP &24.84 (24.66)  \\
    Walklets  &  41.37(41.19)  \\
    Node2Vec (p=q=0.25)   &  26,96 (25.81)   \\
    Node2Vec (p=q=1)  &26,52 \\
  \bottomrule
\end{tabular}
\end{table}

As clearly visible in table \ref{tab:rpsc}, our reproduced results are close to the reported ones and in particular the reproduced results are constantly better. The slight deviation can be explained by the random factor. We can assume that our used parameters are mostly similar to those used in the original papers. 

The different settings of \textit{p} and \textit{q} in Node2Vec show that changes of these parameters affect the scores only slightly. The choice of \textit{p} = \textit{q} = 0.25 leads to a small improvement over \textit{p} =  \textit{q} = 1, but does not have a huge impact. Therefore, it can be assumed that a cost-intensive parameter search regarding \textit{p} and \textit{q}  is not essential. 
At the setting of \textit{p} = \textit{q} = 1, Node2Vec resembles DeepWalk, as in this setting, the random walk sampling is not biased, rendering the walks really random.
The difference between Node2Vec with \textit{p} = \textit{q} = 1 and DeepWalk in the table can be explained by the differences in the parameter choices for walk length \textit{t} and number of walks $\mathbb{\mu}$.
As a result of the minor difference between the different choices in Node2Vec's walk control parameters, we only use DeepWalk in all subsequent tasks.

\subsection{Node Classification}
\label{sssec:nc-results}
We evaluated HALK, HARP, DeepWalk and Walklets on the node classification task with the following datasets: Cora, BlogCatalog and CiteSeer (see table~\ref{tab:fre} for basic statistics).
First, we learned embeddings for all nodes in the graph and then we evaluated the same supervised classifier as in the reproduction experiment over training / test splits.
For Cora and CiteSeer we used 90\% of the data for training and 50\% for BlogCatalog as BlogCatalog contains more vertices and edges.
Similar to the reproduction experiment, we strictly ensured that all methods use the same data for training and testing to make a fair comparison possible. 
We report the Macro-F1 score (and standard deviation) averaged over 10 random splits.


As all compared methods are random walk based, we used the same parameters that determine the characteristics of the  random walks for HARP, HALK, DeepWalk and Walklets. 
We also used the same representation and window size for these methods. 
Most parameters are similar to those that were already used in the reproduction of results. 
In particular, we set $\mathbb{\mu}$ = 80, \textit{t} = 40,\textit{w} = 10,  \textit{d} = 128, negative samples = 5 and sample=0.1 for HALK, HARP, DeepWalk and Walklets. 
We also set $\mathbb{\alpha}$ = 0.025,  $\mathbb{\alpha}_{min}$, = 0.001 and minimum count = 0 for all methods. 
For DeepWalk and Walklets, we set the number of iterations to 5.
These settings mostly follow the original node classification setup of DeepWalk. 
We set the skip window-size for Walklets to $\mathbb{\pi}$ = 2. 
We trained HALK on 4 levels with 10\%, 20\%, 40\% and 100\% of the most frequent nodes and 10, 5, 3, 1 iterations respectively.
While 10 iterations may seem rather high, one needs to consider that at this level, only 10\% of the nodes are used for training, effectively resulting in less training time than a single iteration on the full data.

\begin{table}[tbp]
\center
  \caption{Node classification Macro-F1 scores and standard deviation ($\pm$) over 10 random splits. Best score in bold, second-best underlined.}
  \label{tab:ncr}
  \begin{tabular}{lccc}
    \toprule
    Algorithm/Dataset&Cora&BlogCatalog&CiteSeer\\
		\midrule
    HALK        & \textbf{81.65} $\pm$2.2       & \textbf{27.70} $\pm$0.4       & \textbf{56.97} $\pm$3.2  \\
    HARP        & \underline{81.33} $\pm$2.5    & \underline{27.65} $\pm$0.6    & \underline{56.16} $\pm$2.0  \\
    Walklets    & 81.10 $\pm$2.5                & 27.31 $\pm$0.5                & 55.76 $\pm$2.9  \\
    DeepWalk    & \underline{81.33} $\pm$2.0    & 27.57 $\pm$0.5                & 55.54 $\pm$2.8  \\
  \bottomrule
\end{tabular}
\end{table}

The results are presented in table \ref{tab:ncr}. Best score per dataset is marked in bold, second best is underlined. HALK performed best on all three datasets with our parameter settings. 
However, as differences in scores are all well within the standard deviation, there is practically no difference in performance. 
That means, we do not have a winner, but all methods perform equally well. 

Even more, running the evaluation a second time with the same parameter settings may result in a slightly different ranking.
We observed this behavior for example for DeepWalk, creating several (Deepwalk) embedding models on the same set of random walks. 
Some of these models were then able to outperform the score of HALK on CiteSeer, but had a lower score than reported in the table on Cora.
This behavior can be explained by the random initialization of the embeddings (as the randomness introduced by the random walks is eliminated by using the same set of walks).

%

\subsection{Link Prediction}

Link prediction attempts to estimate the likelihood of the existence of edges among nodes based on the observed network structure. 
For instance, recommendation systems need to predict missing friendship links in social networks and affinities between users and movies.  
In computational biology, interaction graphs (e.g. proteins, drugs or diseases) are usually incomplete and predicting links in these noisy graphs is very important. 
Moreover, link prediction is commonly used for statistical relational learning to predict the relation between entities in a knowledge graph \cite{nickel2016review}.


Given a graph $G=(V, E)$ the link prediction task is to predict the existence of an edge between two nodes. From a pure prediction perspective, of course the task is to predict were edges should exist or connections will be established in the future. 
The positive examples are obtained by removing  $50\% $ of edges from the original graph randomly,
whereas negative examples are generated by randomly sampling an equal number of node pairs that are not connected by an edge (i.e. $ (i,j) \notin E$ ). 
Exploiting the vector representations from each embedding technique, we learn a model to predict whether a given edge in the test set exists in $E$ or not. 
The prediction task involves pairs of nodes, hence we need to use a bootstrapping approach over the feature representations of the individual nodes. 
We use the same set of binary operators as Grover and Leskovec~\cite{Grover2016} to construct feature representations for the edges: L1, L2, Hadamard and Average (see table~\ref{tab:opr} for details).
These edge features are input to a logistic regression classifier with L2 regularization to perform the binary classification task.

\begin{table}[t]
  \caption{Choice of binary operators.}
  \label{tab:opr}
	\center
  \begin{tabular}{lcc}
    \toprule
    Operator &Symbol& definition\\
    \midrule
		L1 & $|\ominus |$ & $|\phi_i(u)-\phi_i(v)|$\\
		L2 & $L_2$ & $|\phi_i(u)-\phi_i(v)|^2$\\
    Subtraction & $\ominus $ & $\phi_i(u)-\phi_i(v)$\\
     Concatenation & $\oplus$&  $(\phi(u),\phi(v))$\\
     Average  &  $\oslash$ & $\frac{\phi_i(u)+\phi_i(v)}{2}$ \\
     Hadamard & $ \odot$  &$\phi_i(u)*\phi_i(v)$\\
  \bottomrule
\end{tabular}
\end{table}
We re-used the embeddings for BlogCatalog obtained in the node classification task and kept all hyperparameter settings the same as in the node classification task for the Facebook dataset.

Table~\ref{tab:reg} illustrates the results of our analysis reported by the score of AUC (Area Under Curve). 
\begin{table}[b]
\caption{Area Under Curve (AUC) scores for link prediction. Best score per dataset in bold, best score per dataset and operator underlined.}

\begin{center}
\begin{tabular}{llcccc}
\toprule
Dataset&
Embedding&
\multicolumn{4}{c}{AUC} \Tstrut \\
\midrule
& &  $\oslash$ & $|\ominus |$ &$L_2$ &$ \odot$   \Tstrut \\ 
\multirow{5}{*}{Facebook}   & HALK      & \underline{0.780} & 0.986 & 0.986 & 0.991 \Tstrut \\ 
                            & HARP      & 0.755 & 0.993 & 0.994 & \underline{0.988} \Tstrut \\ 
                            & WALKLETS  & 0.767 & \underline{0.994} & \underline{\textbf{0.995}} & 0.987 \Tstrut \\ 
                            & DeepWalk  & 0.765 & 0.992 & 0.992 & \underline{0.988} \Tstrut \\ 
\midrule
\multirow{5}{*}{BlogCatalog}    & HALK      & 0.935 & 0.949 & 0.953 & 0.821 \Tstrut \\ 
                                & HARP      & 0.940 & 0.975 & 0.977 & 0.790 \Tstrut \\ 
                                & WALKLETS  & 0.913 & \underline{0.984} & \underline{\textbf{0.985}} & \underline{0.836} \Tstrut \\ 
                                & DeepWalk  & \underline{0.944} & 0.979 & 0.981 & 0.805 \Tstrut \\  
\bottomrule 
\end{tabular}
\label{tab:reg}
\end{center}

\end{table}
Best scores per binary operator are underlined and best overall scores per dataset are marked bold.
It can be seen that performance of link prediction greatly varies depending on the binary operation. 
A general observation we can draw from the results is that L1 and L2 achieve very close performance for every embedding techniques. 
We can confirm the Hadamard operator as ``higly stable''~\cite{Grover2016} only partially. 
While its performance on the Facebook dataset is consistent across all four methods and close to optimum, it varies slightly more on BlogCatalog.
Also on BlogCatalog, the Hadamard Operator performs considerably worse than the other operators, which all yield results in a similar range.
When we look at operators individually, Walklets outperforms the others, both in L1 and L2, which also results in the best overall score.
For the remaining operators, the best performing method varies.
However, similar to node classification in the previous experiment, scores are extremely close to each other and we cannot determine a clear winner.

\subsection{Shortest Path Approximation}

The first observation we can draw from the results is the consistently low performance of the subtraction operator $\ominus$. 
In fact, its performance is equal to random prediction.
That means, combining the node features of a pair to an edge feature via subtraction does not retain any meaningful information in the resulting edge embedding.

On the Facebook dataset, the results per binary operator are rather consistent among the different methods (ignoring random embeddings as an obvious exception).
The only exception is the Hadamard operator, which yields intermediate scores for HARP and DeepWalk, a score at the top end for HALK and one at the lower end for DeepWalk.
On the Youtube dataset, the Hadamard operator behaves even worse, resulting in a score below random for the Walklets approach.
Hence we cannot consider Hadamard as a stable operator for the shortest path approximation task.
While the L1 operator is rather stable and provides the best results on the Facebook dataset, its performance varies stronger on the Youtube dataset and drops to the lower end.

In terms of absolute values, HALK performs best on Facebook and DeepWalk and Walklets share the top score on Youtube (on which HALK performs worst).
However it is again hard to draw a decisive conclusion and select a clear winner, given the variance across datasets and operators and the (partially) small to non-existent differences in the achieved scores.

All methods clearly outperform the trivial and random baseline in terms of MAE on both datasets. 
On Facebook, trivial and random baseline are on par, whereas on Youtube, the random baseline performs even worse than the trivial predictor.
It seems as if the linear regression has been fooled, learning in-existent patterns.
In terms of MRE all embedding methods improve only slightly over the baseline.
This minor improvement is caused to a partial extent by the nature of the MRE as mentioned before.
However, the improvement in terms of MAE is also smaller when compared to the Facebook dataset, so the behavior is not only explained by the MRE's nature.
The second part of the explanation is the imbalance of the walk length: In the test set, the difference in shortest path frequencies on Facebook is between a 6-digits number for the most and a 4-digits number for the least. On Youtube this difference is by far more pronounced with a 7-digits number for the most and 2-digits for the least.


\section{Conclusion \& Future Work}
\label{sec:conclusion}
We conclude, that the embedding methods tailored towards retaining long distance relationships or representing the graph at different hierarchical (including our method) can improve the performance under certain circumstances, such as parameter settings, particular tasks or datasets, but in general the difference in performance is neglible.
DeepWalk is still along the state of the art and from the methods compared in the evaluation in this paper, it is the simplest method.
That means while a more complex method might outperform DeepWalk on a certain task/dataset, due to its simplicity it is preferable in a general setting.
We were surprised by the similarity of results obtained when conducting a rigorous evaluation.
Even though all methods used for comparison claim improvement over DeepWalk, we did not clearly see this reflected in our evaluation.
When selecting parameters that are known to provide top scores, differences almost vanish.

We plan to conduct a large-scale evaluation, including further methods and datasets, since an unbiased comparison of different methods is strongly desirable. 
The reproduction of results in section~\ref{ssec:reproduction} explicitly shows the need for such a comparison: 
While the evaluation setups are highly similar across the compared methods, reported results do not match the best known results. 



\bibliographystyle{IEEEtran}
\bibliography{halk}

\end{document}